\documentclass{amsart}
\usepackage{xcolor}
\usepackage{graphicx}
\usepackage{cite}
\usepackage{url}

\usepackage{amsmath}
\usepackage{amsthm}
\theoremstyle{definition}
\newtheorem*{definition}{Definition}

\begin{document}
\title[The cost of passing]{The cost of passing -- using deep learning AIs to expand our understanding of the ancient game of Go}

\author{Attila Egri-Nagy$^1$, Antti T\"orm\"anen$^2$}
\address{$^1$Akita International University\\Department of Mathematics and Natural Sciences\\ Yuwa, Akita-City 010-1292, Japan}
\address{$^2$Nihon Ki-in -- Japan Go Association\\7-2 Gobancho, Chiyoda City,\\ Tokyo 102-0076, Japan}
\email{egri-nagy@aiu.ac.jp,\ tormanen.antti@gmail.com}

\maketitle

\begin{abstract}
  AI engines utilizing deep learning neural networks provide excellent tools for analyzing traditional board games.
  Here we are interested in gaining new insights into the ancient game of Go.
  For that purpose, we need to define new numerical measures based on the raw output of the engines.
  In this paper, we develop a numerical tool for automated move-by-move performance evaluation in a context-sensitive manner and for recognizing game features.
  We measure the urgency of a move by the cost of passing, which is the score value difference between the current configuration of stones and after a hypothetical pass in the same board position.
  Here we investigate the properties of this measure and describe some applications.
\end{abstract}

\section{Introduction}
AlphaGo \cite{AlphaGo2016} made history by being the first Go program capable of winning against a top human professional player.
The event also changed the goals of artificial intelligence research projects.
Producing ever stronger engines is still a worthy pursuit since the game is not solved yet.
Current research in that direction works in the deep learning paradigm and tries
to optimize the process by changing neural network architecture, the activation
function, or the training algorithm (see e.g., \cite{Cazenave2021}).
However, there are now new possible applications.
Superhuman Go engines can be used to deepen human understanding of the game \cite{shin2021human,proceedings2022081022}.
Here we develop tools for context-sensitive evaluation of the value of moves and automated feature detection of games.

Traditionally, we divide a game of Go into three main parts: the opening, the middle game, and the endgame.
These all have different characteristics, and we apply different types of thinking and skills in each stage.
Of course, these distinctions may not be apparent in a particular game.
For instance,  the stages can overlap. 
A middle-game fight can erupt in a corner and reach its conclusion, while the other corners remain in the opening stage. 
Still, the stages provide a natural framework for understanding games.

Here we augment the traditional division of the stages of the game with a quantitative and fine-grained measure. 
We observe that different stages of a game have different costs for mistakes. 
For example, multiple inaccuracies in the opening can be compensated by the opponent's single larger mistake during a middle-game fight.
It seems that we can characterize board positions by the cost of a mistake.
However, a mistake is context-dependent, and so the question arises: how can we define a bad move precisely and independently from the situation?
In Go, every move is an investment towards some end. In the vast majority of game situations, playing any stone is better than not, and therefore, passing a move can represent a maximally bad move.
Passing during the game is not universally bad: for example, when a game is
over, playing a stone inside one's own or the opponent's territory costs a
point, while passing avoids losing a point.
Also, in general, there can be bigger mistakes on the board (such as killing one's own
group) than passing itself.
However, for game analysis purposes, the cost of passing is a natural and practical tool helping to assess performance move-by-move.

In this paper, first we will give a definition for the cost of passing, then we
describe the core idea for implementing its calculation.
We also give a general description of how the cost of passing changes in
different stages and in different situations of a game.
We separately discuss the concept of sente (having the initiative) in terms of
the cost of passing.
We briefly mention a few possible applications of this measure and describe our
current methods.
Finally, we conclude with the plan of future research based on the cost of passing.

\section{Defining the cost of passing}

We denote the board position after $i$ moves by $s_i$, and $s_0$ refers to the empty
board.
We use function notation for moves: $a_{i+1}(s_i)=s_{i+1}$, i.e.~the $i+1$th
move produces the $i+1$th board position.
In particular, $a_1(s_0)=s_1$, the first move produces the second board
position, indexed by 1, as for stone configurations on the board, the indices always refer to the number of moves made.

For a given board position, deep learning Go-playing AIs can provide information about the probabilities for
winning, the \emph{win rate}, $V(s)$, and the final result score estimate, the score lead,
or \emph{score mean} $m_s$.
These are statistical measures, and due to the probabilistic nature of the tree
search, they are subject to random fluctuations.
As a convention, we always compute the score mean from the perspective of the
black player.
Thus, a positive score mean predicts a black win, while negative values show white's
advantage.

The effect of a move is the difference between the score lead estimate before
and after the move \cite{egrinagy2020derived}.
This describes the efficiency of a move compared to the AI's best move candidate. 
However, board positions and their available moves can have very different characteristics: in some board positions there is only a single good move available, while in other positions dozens of moves can be similarly good. Furthermore, mistakes can be more costly in some board positions than others.
The objective value of a move therefore depends on the game context, and it would be beneficial to have a way to measure the value of a move while taking the whole-board situation into account.

\begin{definition}[Informal]
The \emph{cost of passing} in a board position for the player in turn is the
difference between the score means before making a move and after passing.
  \end{definition}
  In other words, the cost of passing is the price to pay for a missed move.

\begin{definition}
$c_i=\mu_{s_i}-\mu_{\text{pass}(s_i)}$
\end{definition}
As we adopted the convention to report the score mean always
from the perspective of the black player, we need to change the sign for the
white player.
This way, the cost of passing is in general a positive value.
The main exceptions to this are when a game is finished and neutral points are being filled out, during when the cost of passing will be zero, and when a game is completely finished, during when passing in fact saves a point. It should be noted that in the former case, the probabilistic nature of the AI analysis might in fact also output minuscule negative values.

Technically, we could define cost of passing in terms of the win rate.
However, it is more informative to evaluate moves in terms of expected scores.
It is a single bit information to know who won the game, but it is lot more
informative to
know by how much.

\section{Implementation}

For playing the game, the AI engines do not need to calculate the cost of passing.
They evaluate board positions in terms of winning probabilities and estimated scores.
These are already derived measures, as choosing a move is usually decided by the
visit count of the corresponding board position.
The engine chooses the move most often considered by the Monte-Carlo tree search.
Therefore, we need to calculate the cost of passing externally.

The idea is simple: we carry out a hypothetical pass move.
We gain useful information (about an actual board position) by thinking about an
event (pass) that has not happened.
This is a form of \emph{counterfactual reasoning}, which is a crucial tool in causal
inference \cite{pearl2018book}.
Counterfactuals are considered to be important for building future AI systems,
and here they are already important in current applications.

Let's say we want to find the cost of passing for black. We evaluate the board position and record the score mean value.
Then, we leave the same configuration of the board but evaluate it as if it was white's turn.
This direct manipulation of the turn may not work well with the engine, as they take into account the history leading to the board position (to avoid repeating an earlier board configuration).
Thus we need to add a pass to the sequence of moves explicitly.
The cost of passing is the difference between the two score estimates.
This simple trick works for all engines that can do score estimates.

The drawback of this method is that we need to double the number of evaluations.
Game analyses are computation-intensive and consequently time-consuming.
Therefore we have a trade-off: the cost of passing gives more information but requires more resources.
In practice, one needs to evaluate the required strength of the analysis and set the number of visits per move accordingly.

\section{The dynamics of cost of passing in top-level games}

For a given game, we calculate the value of the cost of passing for each board position.
We claim that this sequence of values contains descriptive information about the game.
Here we describe how the values and their changes can characterize the different situations in the game.

\subsection{Linear descent - the background dynamics}

\begin{figure*}[h]
  \includegraphics[width=\textwidth]{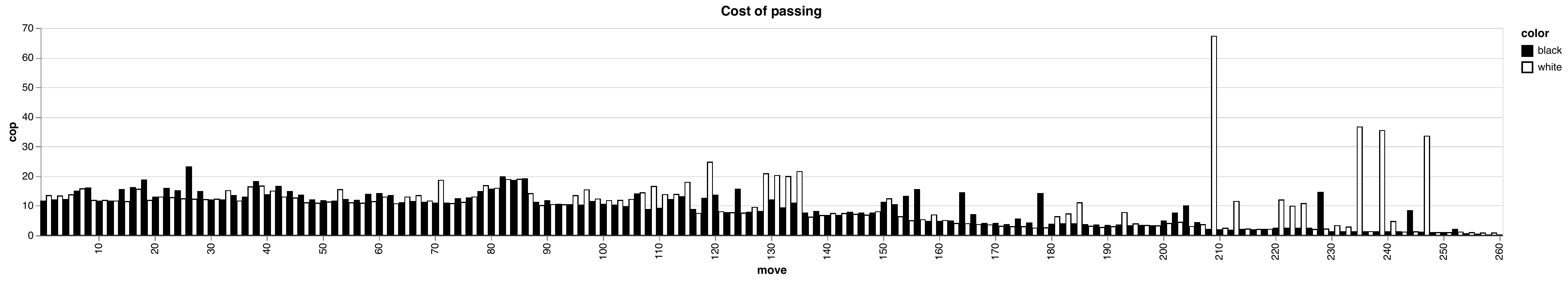}
  \caption{The cost of passing for a relatively peaceful professional game (see
    Game 1 in the Appendix for the game record). The
    linear descent can be clearly seen. There are still individual forcing moves
    (large spikes), one busy exchange for both players (at around 75–88), and
    one one-sided forcing sequence (130–136). The total cost of passing for the
    216 moves is 2516, averaging to 9.64 per move.}
  \label{G1cop}
\end{figure*}

A game of Go finishes when neither player can change the size of territories on the board.
Two consecutive passes end the game.
Thus, the cost of passing is zero when a game is finished.

Traditionally, the value of a single handicap stone has been estimated at roughly 10 points. Giving the black player an extra handicap stone is identical to the white player passing an extra time, so we can use the cost of passing to approximate numerical values for handicap stones. On an empty board, this value comes to roughly 12 points (see the methods section for the details of the engine and the network).

The game starts from the empty board and proceeds to a `crystallized' state, the final configuration.
On the one hand, in the early stages of a game there are more open possibilities
than towards the end; and on the other, because groups on the board need to
secure two eyes in order to survive, the fewer there are stones on the board,
the larger the effect that a single stone has on its surroundings. Therefore, in
general, the cost of passing follows a linear descent.
This theoretical line can be clearly visible from the cost of passing graphs
resulting from AI analysis (see Fig.\ \ref{G1cop}).

Parts of the game, however, deviate from the linear descent: these are when `forcing moves' are played. A forcing move is a move that threatens to make a larger gain if the opponent does not respond. As far as the rules of Go are concerned, a player is always free to play on any open intersection on the board; but if a player wants to win the game, there are situations when it is strictly necessary to respond to the opponent's threat. For the side that plays the forcing move, it can be strategically beneficial to force the opponent to respond in a particular way.
When the whole-board situation stabilises and no forcing moves are being played, the cost-of-passing graph returns to the linear descent, which is its fundamental shape.

All games of Go do not follow quite the same linear descent graph. This is because games have variable lengths, and the length of stages of games also differ. Some fighting-oriented games may only enter the endgame after move 200, in which case the cost of passing remains high for a long time. Some peaceful games may enter the endgame before move 100, after which the cost of passing may stay at below seven points.
A player can resign any time in the game in case it becomes decidedly one sided.
Obviously, winning by resignation implies that at the end of game it is possible
to have cost of passing values significantly different from zero.



\subsection{Cost of passing as `temperature'}

The idea of a measure that is high in the beginning and drops to zero by the end of the game is not new.
This is often called `temperature,' roughly described as the urgency of making a move and often described as the value of the biggest possible move on the board.
The term can be used with mathematical precision in combinatorial game theory \cite{berlekamp2001winning,berlekamp1994mathematical} or in a more intuitive general sense.
The cost of passing may simply seem like a way to calculate this temperature precisely, but there are some decisive differences.
In Go, we often distinguish between the local and ambient temperature, while the cost of passing is calculated for the whole board.
Due to the local versus ambient distinction, the temperature only goes down as the game progresses, while the cost of passing can also increase.
Therefore, it is justified to keep the two concepts separate.

\subsection{The stages of the game characterized numerically}

Generally, as can be judged from the AI's analyses, the `baseline' cost of passing in the early game ranges from about 13 to 10 points. In the middle game, values are generally between 12 and 7 points, and values below 7 points usually indicate the endgame.

\subsection{The `heated' parts: fighting sequences and threats}
\begin{figure*}[h]
  \includegraphics[width=\textwidth]{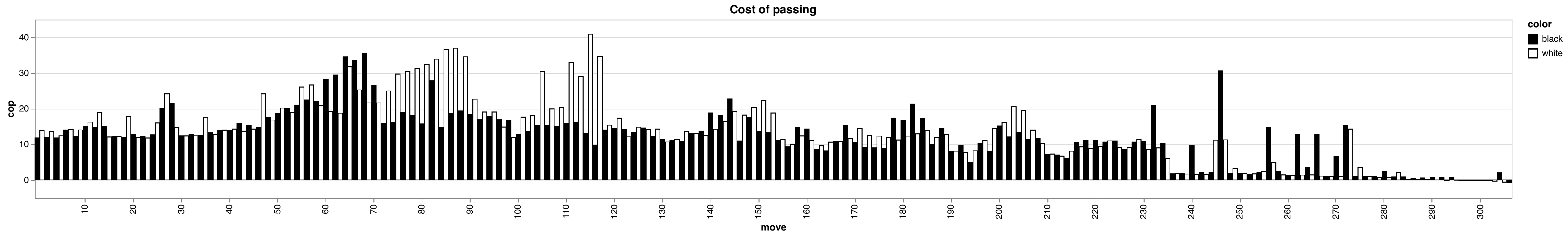}
  \caption{AI versus AI game, KataGo playing itself but with different networks
    (40-block versus 60-block). It is a fighting game indicated by the elevated
    level of cost of passing. The one-sided black and two subsequent white peaks
    show white and then the black player dominating the fight
    (see
    Game 2 in the Appendix for the game record). The total cost of passing for the
    307 moves is 3937.93, averaging to 12.83 per move.}
\label{G2cop}
\end{figure*}

The cost of passing can detect different game situations based on the deviations
from the baseline linear descent.
Fighting sequences are characterized by values that are high for
both players.
Threats can be recognized when the elevation is one-sided, where it is high only for the defending player.
See Fig.\ \ref{G2cop} for a fighting game exhibiting continuously elevated
levels of cost of passing.

\section{On sente}
Sente, or having the initiative, is a fundamental concept in the game.
It can roughly be translated into having the freedom of deciding where to play.
The opposite is gote, having no choice but to make a forced move to avoid a larger
loss, in effect merely reacting to the opponent's actions.
Often but not always, the player with sente has the strategical upper hand in a game.

\subsection{On detecting sente}

\begin{figure}[h]
  \includegraphics[width=.5\textwidth]{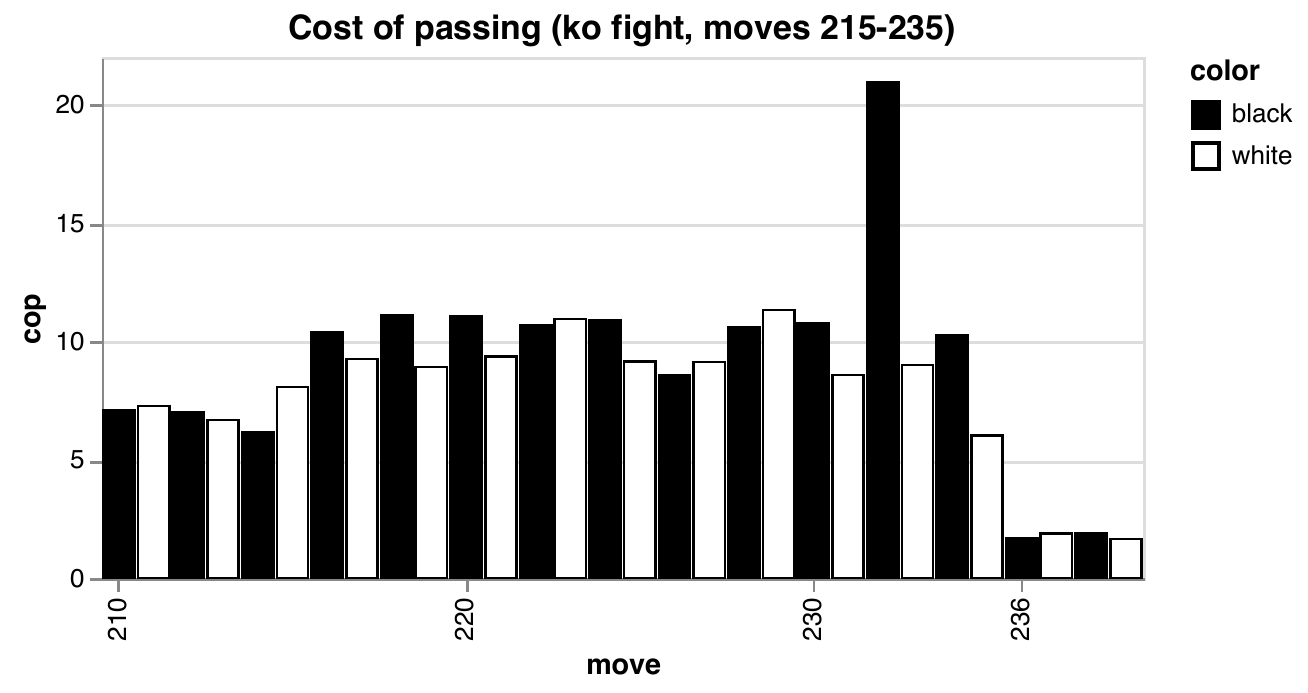}
  \caption{Black provokes a fight that develops into a ko fight, hence the
    elevated cost of passing for the sequence. White wins the ko, but the ko
    needs to be finished in gote, thus after 236 moves, Black has a clear sente
    move. (see
    Game 2 in the Appendix for the game record)}
  \label{G2kofight}
\end{figure}

The cost of passing makes it possible to define mathematically whether a player
has sente or not.
When the cost of passing is elevated beyond the baseline suggested by the linear
descent graph, there is something `urgent' on the board and a player is not free
to choose their action; in this case, the player has gote.
If the cost of passing is instead at the baseline, then there are multiple
similarly good moves available, and the player's action is not `forced'; in this
case, the player has sente.
When a fight is finished, there is a drop from an elevated value to a baseline
value of the cost of passing, and the player who has the baseline value has sente.
See Fig.\ \ref{G2kofight} for an example.

\subsection{On the value of having sente}

When professional players estimate the score in a game, they usually count the values of secure territories and then weigh unfinished, potential territories against each other. Finally, to finish this process, the player has to take into account whose turn it is, as having the move turn (i.e., sente) is also worth something.

The value of having sente is generally half of the cost of passing. This is mathematically self-evident: if a player passes, they give the sente to the opponent. The player first had whatever was the value of having the sente, and after the pass it is the opponent who gets the value, so the total difference is twice the value.

At the start of the game, the value of sente is equal to correct komi, or the white player's compensation for the fact that black goes first. Under Japanese rules of Go, this is currently 6.5 points.

For example, a professional player might estimate that, in a given game position, black has 40 points and white 30 points of secure territory and the players' territorial potentials are roughly equal. If the game has just entered the endgame, the cost of passing might be about seven points. If it is white's turn, half of seven points and 6.5 from the komi have to be added to his score, so the estimation suggests that white is roughly one point ahead of black.




\section{Applications}

Here we sketch a few possible applications of the cost of passing.
Our focus is on the human side, i.e.~we would like to facilitate the learning
process for human players.

\subsection{Danger level indicator}

Beginning players often have issues with recognizing immediate danger.
This can easily lead to a lost game, even if the player would have known how to
defend.
As a direct solution, we can implement the advice directly in a teaching software application by highlighting the critical area and possibly the required defensive moves.
Some Go clients have the option to warn about atari, as a basic and very special
case of this idea.
However, it could be more beneficial for the learner to merely indicate the
presence of the danger, but not the location, and let her find out what to do.

The cost of passing is suitable for implementing such a danger level indicator.
The higher the price for passing a move, the higher the level of danger.
Seeing the warning, the player still has to scan the board for the threat, which can lead to good
playing habits.

\subsection{Automated selection of points of interests}

When working with a large collection of game records we may want to find many
examples of certain situations like fighting sequences, or threats and defences.
One obvious way to do this is to have an expert player going through game
records.
This time-consuming process can be made more efficient by automatically picking
the points of interest.
However, recognizing the interesting events is difficult by using the existing measures.
Win rate can detect decisive blunders in balanced games, but once a game is one-sided, further blunders go unnoticed.
Using score mean is better and  with the derived measure of effect we can have a move-by-move
analysis of performance.
Still, the effect does not give information about the context of a move.
The cost of passing is designed to give the contextual information.
As we saw, it can recognize features of the game.
Therefore, adding it to the analysis can improve cataloging and
data mining on game record databases.

Possibly the biggest advantage of deep learning AIs over humans is
the speed that they can learn from previous games, therefore automating the task
of finding the points of interest in a game record can improve the human
learning process.

\subsection{Game quality quantifier}

Just the effect of moves (i.e., counting differences between chosen moves and the AI's top candidates) is not a sufficient indicator to calculate the quality of a game. This is because the nature of a game – how peaceful or fighting-oriented it is – directly affects the result. Fighting-oriented professional games can have average effects of -0.6 points a move or more, while peaceful games by intermediate-level amateurs can have average effects of only -0.4 points a move. In reality, however, there will still be a huge level gap between the players of these games.

Instead of merely looking at the effects of moves, it can be more accurate to quantify how big of a percentage of a player's cost of passing throughout a game was `realised' through their moves. If a player's cumulative cost of passing in a game was for example 1,000 points, and the realised value of the player's moves was 950 points, this gives a performance value of 95 percent. This method is strictly fairer than looking at the effects of moves, since the cumulative cost of passing is adjusted for the `complexity' of the game.

More precise profiling of a player's performance is important in developing
tools for cheat-detection.
Cost of passing can help by making the existing tools more efficient \cite{egrinagy2020derived}.

\section{Methods}

We have three main software components in this project:
\begin{enumerate}
\item a deep learning Go engine,
\item data processing and
 \item interactive visualization tools.
\end{enumerate}

We used the KataGo engine for doing the game analysis.
It is a leading open-source implementation of the self-play-trained Go engine
model introduced by AlphaGo Zero \cite{AlphaGoZero2017}.
It incorporates many improvements \cite{KataGo2019}.
Establishing the strength of an engine precisely is difficult since it depends on many factors, e.g., the game settings and the underlying hardware.
Rankings of Go engines are often heavily debated.
Therefore our choice is based on secondary pieces of evidence.
KataGo is used in many analysis tools and serves as a base for several engines in computer Go competitions.
We used version v1.11.0 with networks kata1-b40c256-s11101799168-d2715431527 and
kata1-b60c320-s6321537280-d2951683615.
These are available at \url{https://katagotraining.org/}.

We used a purpose-built LambdaGo software package written in Clojure \cite{Clojure2020}
to process the output of the analysis engine (available at \url{https://github.com/egri-nagy/lambdago}).
This data-oriented general purpose language is well-suited for exploratory
analysis of engine output and preparing visualizations of the data.
For creating the graphs, we used Vega and Vega-Lite visualization grammars \cite{2017-vega-lite}.

\section{Conclusion}
Here we made the next step in the utilization of Go-playing AI engines for improving human understanding of the game.
With the development of the cost of passing measure, we could provide a graph of the
complete game that serves as a visual summary of the events.
It clearly indicates the different stages of the game and the different
types of plays (e.g.~critical moves, threats and defences) that cannot be picked
up by the winrate and the score mean graphs.

We expect that the existing analysis software tools will adopt cost of passing to provide
more information for the human learners.
Also, as a next step we plan a more thorough bulk analysis of
historical and current games, similar to the analysis \cite{ELFanalysis} done by
ELF OpenGo \cite{ELF2019},
but with the cost of passing included.

\textbf{Acknowledgements.} This project was funded by the Kakenhi grant
22K00015, titled `On progressing human understanding in the shadow of superhuman
deep learning artificial intelligence entities' (Grant-in-Aid for Scientific
Research type C, \url{https://kaken.nii.ac.jp/grant/KAKENHI-PROJECT-22K00015/}).

\bibliographystyle{IEEEtran}
\bibliography{IEEEabrv,../compgo}


\appendix

\section{Game Records}

\subsection*{Game 1}

Antti Törmänen 1p versus Shuto Shun 8p, 2022.03.14, W+2.5

\url{https://github.com/egri-nagy/lambdago/blob/master/resources/SGF/2022_cop/antti-shuto.sgf}

\begin{center}\includegraphics{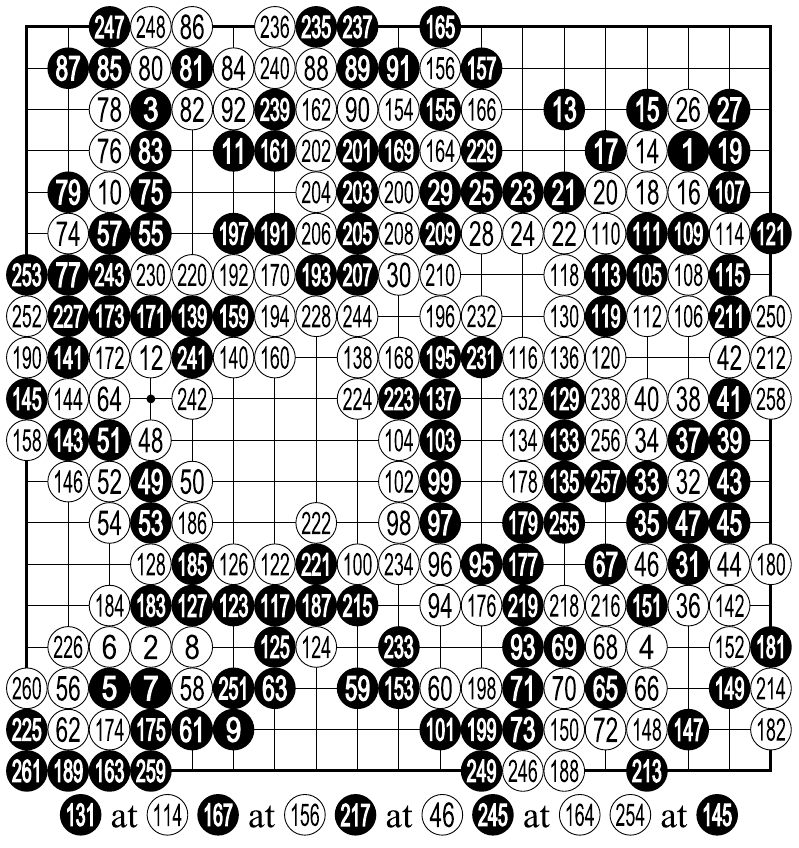}\end{center}

\subsection*{Game 2}

kata1-b40c256-s11101799168-d2715431527 versus
kata1-b60c320-s6321537280-d2951683615 using KataGo v1.11.0

\url{https://github.com/egri-nagy/lambdago/blob/master/resources/SGF/2022_cop/20220810_40bvs60b.sgf}

\begin{center}\includegraphics{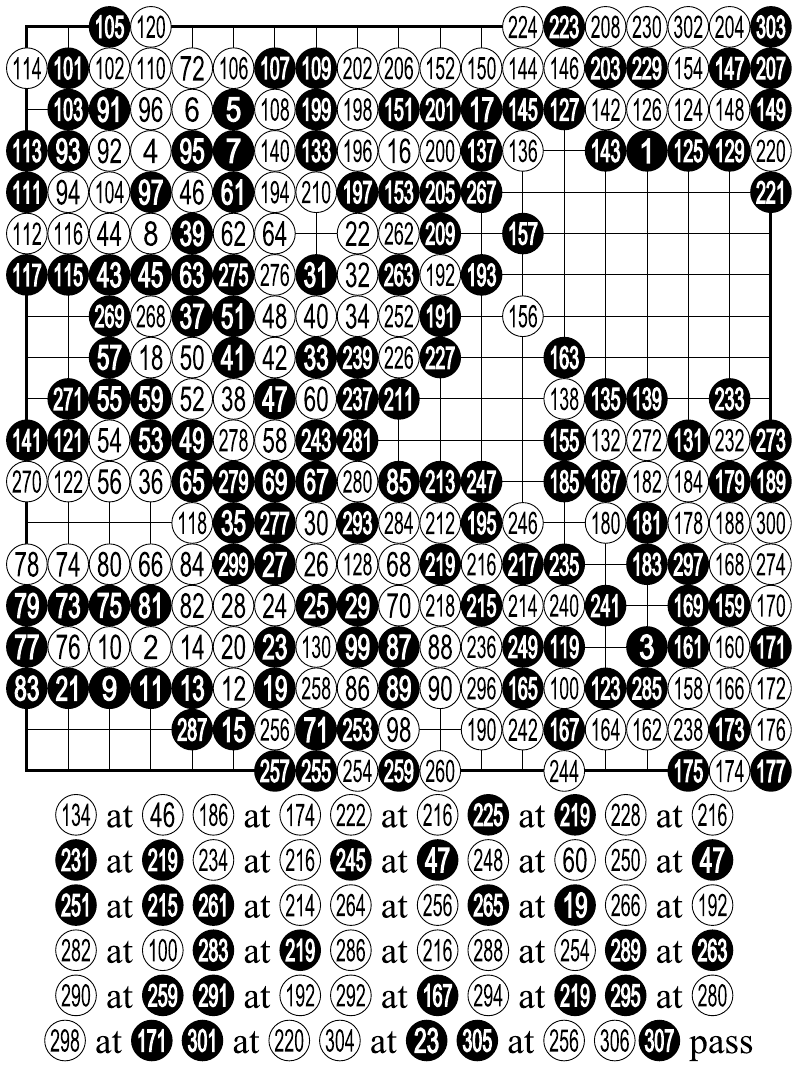}\end{center}

\end{document}